\documentclass[11pt]{article}
\usepackage[preprint]{acl}
\usepackage{times}
\usepackage{latexsym}
\usepackage[T1]{fontenc}
\usepackage[utf8]{inputenc}
\usepackage{microtype}
\usepackage{inconsolata}
\usepackage{graphicx}
\usepackage{amsmath}
\usepackage{booktabs}

\title{Forking Fast: Efficiently Estimating \\ Uncertainty Dynamics in
Text Generation}   

\author{Eric Bigelow, Amir Zur, Satchel Grant, Tal Haklay, Can Rager, Owen Lewis, \\ \textbf{Thomas McGrath, Jack Merullo, Ekdeep Singh Lubana, Atticus Geiger} \vspace{4pt} \\
   {\includegraphics[height=4ex]{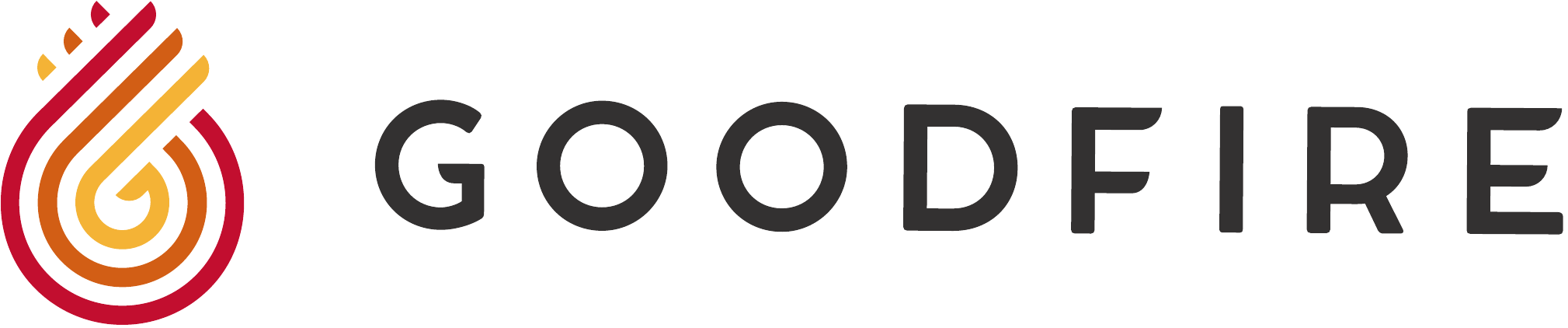}}
  }

\begin{document}
\maketitle

\begin{abstract}
LLM reasoning is stochastic, and so understanding a model requires grappling with the distribution of reasoning chains that it might produce for a given question, i.e., its uncertainty. 
Resampling-based analyses characterize this distribution, revealing which steps of a rollout determine how the model arrives at its answer.
However, a major limitation of these approaches is that resampling text sequences at every token or sentence in a reasoning chain is very costly.
Our work strives to make resampling analysis more computationally efficient, while also shedding light on an important scientific question: what is the right statistical model for explaining uncertainty dynamics in text generation?
We show that when resampling many reasoning chains, uncertainty dynamics converge to stable patterns, and noise is largely an artifact of sampling rather than an LLM's sensitivity to each individual token or reasoning step.
We develop a statistical model for smoothing noisy low-sample rollout data to better approximate high-sample data, allowing us to significantly cut sampling costs.
\end{abstract}

\section{Introduction}
\label{sec:intro}

\let\thefootnote\relax\footnotetext{Code and interactive dashboard: \url{https://github.com/ericb-goodfire/forking-fast}}

When Large Language Models (LLMs) engage in long-form text generation such as reasoning, there are many possible paths they might take at each step or token that they generate. Prior work has shown that specific steps during reasoning can have a significant impact on the model's overall behavior, while many other steps are not as important \citep{bigelow2024forking, bogdan2025thought, zhang2025reasoning, macar2026thought, boppana2026reasoning}. 
One method for understanding which steps of reasoning (and text generation more generally) are most important is resampling: given a single reasoning rollout, resample alternate rollouts at each step of reasoning, then collect the final answers of all rollouts and aggregate them into a distribution.


Resampling methods such as Forking Paths Analysis (FPA) \citep{bigelow2024forking} can provide critical insight into uncertainty dynamics in text generation, where uncertainty measures how likely an LLM is to pick one answer over another. However, this approach is very costly and often requires millions of tokens to analyze a single reasoning chain.
In this work, we propose a more efficient way to approximate these uncertainty dynamics by developing a statistical model of their distribution, and using this model to smooth noisy low-sample estimates.

We evaluate our approach by studying Chain-of-Thought reasoning~\citep{kojima2022large} in Llama-3-8B-Instruct~\citep{grattafiori2024llama} and the native reasoning model DeepSeek-R1-Distill-Llama-8B \citep{deepseek2025r1} as they solve problems in tinyMMLU
\citep{polo2024tinybenchmarks,hendrycks2021measuring}. After collecting and analyzing reasoning data with nearly two billion tokens in total, we find three main results:

\begin{enumerate}
    \item When resampling a small number of times, the uncertainty dynamics of reasoning rollouts are quite noisy~\cite{bigelow2024forking}. However, when resampling hundreds of times, the uncertainty dynamics become increasingly smooth, except at key forking points which have sharp changes. We find that variation across reasoning rollouts is modeled well as multinomial sampling noise (\S\ref{sec:results-noise}), and reconstruction error decays in proportion to the square root of the amount of samples collected.

    \item We develop a statistical model which allows us to approximate high-sample uncertainty dynamics by smoothing lower-sample data. This model uses change point detection to identify forking points, and kernel-pooling to smooth estimates between these points. By smoothing low-sample data, we can effectively multiply the effective sample size by $3.3\times$ ($S=30$) to $5\times$ ($S=5$) (\S\ref{sec:results-recovery}).
    
    \item We find that resampling every $N$ tokens or steps, instead of every step, can improve data efficiency without smoothing. However, this approach loses precision in estimating forking points, and it benefits less from smoothing compared to $N=1$. By combining this approach with our smoothing model, we can cut the total budget by $1/8$ with only a small increase in error (\S\ref{sec:results-power}).
\end{enumerate}

\begin{figure}[t]
\centering
\includegraphics[width=\columnwidth]{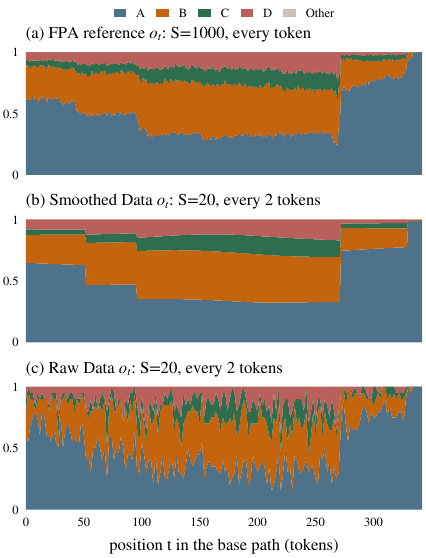}

\caption{Uncertainty dynamics $o_t$ for one question (tinyMMLU question~39; Llama-3-8B-Instruct). Colors represent the fraction of rollouts at $t$ that end with different final answers. (a)~When collecting a massive number of resampled rollouts ($S{=}1000$) at every token ($N{=}1$), the outcome distribution becomes increasingly smooth, except at forking points. (c)~Uncertainty dynamics with a lower-sample analysis ($S{=}20$, $N{=}2$) using $1\%$ of the reference's sampling cost. (b)~Smoothing the lower-sample data $o_t$ closely approximates the high-sample data and recovers the same forking points. More examples in Appendix~\ref{app:more-recons} and our interactive dashboard (\url{https://github.com/ericb-goodfire/forking-fast}).}
\label{fig:reconstruction}
\end{figure}

\section{Methods}
\label{sec:methods}

\paragraph{Forking Paths Analysis}
Following \citet{bigelow2024forking}, we analyze uncertainty dynamics relative to fixed reasoning chain $x$, a \textit{base path}, generated by greedy decoding. At each token or sentence position $t$ per interval $N$, for example every token or every sentence ($N{=}1$), we resample continuations from the prefix $x_{<t}$. Specifically, for every alternative token with probability $p(x_t \mid x_{<t}) \ge 0.05$, we sample $S$ continuations $x_{>t}$ with temperature $\tau = 1.0$ and extract each continuation's outcome, in our case a multiple choice answer \textit{A/B/C/D/Other}. We then aggregate these final answers into a weighted distribution over outcomes, such that each timestep $t$ indexes a distribution $o_t$ over final answers weighted by the probability of the token $p(x_t \mid x_{<t})$ and continuation $p(x_{>t} \mid x_{\leq t})$, which can be visualized as a timeseries (Fig.~\ref{fig:reconstruction}). 

In this work, we consider two methods for reducing the sampling cost of FPA: decreasing the number of samples $S$ at each token position, and increasing the sampling interval to resample every $N$ tokens or sentences, instead of at every step (i.e. $N{=}1$). Our goal is to use this lower-sample data to approximate uncertainty dynamics in $o_t$ with high $S$ and $N{=}1$.

\paragraph{Estimating uncertainty dynamics}
When do sharp changes in the outcome distribution $o_t$ correspond to decision points in the reasoning path, and when do they correspond to sampling noise?
Figure~\ref{fig:reconstruction}a shows a reference $o_t$ with a high resample rate $S{=}1000$ and $N{=}1$.
We find that with large amounts of resampling data, the outcome distribution $o_t$ becomes increasingly smooth at almost all positions $t$, except for key \textit{forking points} where $o_t$ suddenly and dramatically changes \citep{bigelow2024forking}.
Compare this to Figure~\ref{fig:reconstruction}c, which shows a lower-sample outcome distribution $o_t$ with $S{=}20$ and $N{=}2$.
It is difficult to visually discern from the lower-sample data whether sharp changes in $o_t$ are genuine uncertainty dynamics, or whether these fluctuations are an artifact of sampling noise.

Since we draw $S$ continuations independently for each possible next-token $x_t = w$, the answer counts form a multinomial distribution $c_t^{(w)} \sim \mathrm{Multinomial}(S, o_t^{(w)})$, and the branch weights $\tilde{p}_w$ are known exactly from the next-token distribution $p(x_t = w \mid x_{<t})$, so the weighted estimate of $o_t$ has variance $\sum_w \tilde{p}_w^{2}\, o^{(w)}_{t,k}(1-o^{(w)}_{t,k})/S$ for each answer category $k$.
Sampling a branch weighted by $\tilde{p}_w$ and one of its continuations yields i.i.d.\ draws from $o_t$ itself, so counts constructed this way satisfy $c_t \sim \mathrm{Multinomial}(S, o_t)$ exactly. Therefore, we expect sampling noise to decrease in proportion to $1/\sqrt{S}$. That is, we expect a negative linear relation between the log distance between the true outcome distribution (which we approximate with high $S$ and $N{=}1$) and $\sqrt{S}$ as we increase the number of resampled outputs. We empirically test this in Section~\ref{sec:results-noise}.




\paragraph{Statistical model}
If we simply increase the sample count $S$, we can expect to get closer to the model's true uncertainty dynamics $o_t$ at a rate of $\sqrt{S}$. Next, we propose a method for more efficiently approximating the true $o_t$ using reduced sampling rates by varying $S$ and $N$.

Our statistical model builds on a few qualitative observations about the shape of $o_t$. We observe, as do \citet{bigelow2024forking}, that uncertainty dynamics in $o_t$ follow a pattern where large segments of $o_t$ are relatively stable, remaining flat or very slowly drifting over $t$. These stable segments are separated by sharp changes at \textit{forking points}, where the outcome distribution suddenly changes, sometimes after only a single token.

Our model follows a three-stage process: change point segmentation, kernel-weighted Dirichlet pooling, and cross-validation tuning.
First, we apply pruned exact linear time (PELT) change point detection \citep{killick2012optimal,truong2020selective} with an exact multinomial cost. These change points divide the base path into segments in which we expect the outcome distribution to either remain flat or slowly drift. Next, within these segments we pool answer counts from neighboring data points, weighting them with a Gaussian kernel, and use the result to parameterize a Dirichlet distribution.
Finally, we use cross-validation to tune three hyperparameters of the model: the cost function (multinomial likelihood vs. $L_2$ loss) and penalty value for PELT, and the kernel bandwidth used for Gaussian smoothing within segments. This cross-validation operates over the same data being smoothed, by splitting e.g. $S{=}20$ into 5 folds each with $S{=}4$ data points and finding the hyperparameters for each $S{=}16$ that best explains the held out fold.

We evaluate the overall success of our statistical model by measuring total variation distance (TVD) of smoothed low-sample $o_t$ relative to ground truth high-sample $o_t$ data with $S{=}200, N{=}1$.
TVD measures the distance between two distributions $p$ and $q$: $\text{TVD}(p, q) = \frac{1}{2} \sum_{k=1}^K \mid p_k - q_k \mid$.
In Appendix~\ref{app:ablation} we report ablations to our model, which confirm that PELT can accurately identify forking points and that kernel pooling produces accurate flat regions.

\begin{figure}[t]
\centering
\includegraphics[width=\columnwidth]{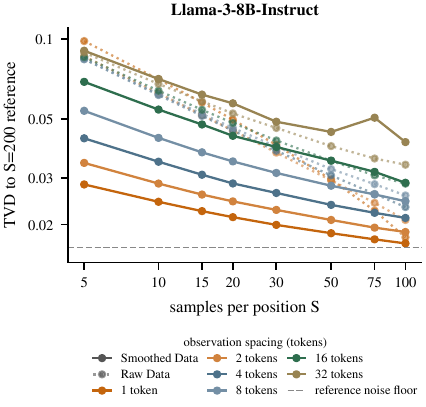}
\caption{\textbf{Accuracy versus number of resampled rollouts.} \ \ 
We find that with raw data observations $o_t$ (dotted lines), greater resample spacing $N$ leads to slightly lower accuracy with equal $S$, with considerably more difference at higher $S$. This effect is significantly more pronounced with the smoothed $o_t$ (solid lines), which also shows that increasing $\log S$ has a consistent linear effect on log-TVD.
Larger $N$ having decreasing accuracy for a given $S$ has a relatively simple explanation, since this means fewer samples collected overall.
Accuracy on the y-axis is measured by total variation distance (TVD) relative to the $S={200}$ reference $o_t$.
}
\label{fig:accuracy}
\end{figure}

\section{Experimental Results}
\label{sec:results}

\paragraph{Data.}
We analyze forking paths on the tinyMMLU dataset~\citep{polo2024tinybenchmarks} ($n{=}100$), collecting a high-sample reference FPA dataset with $S{=}200, N{=}1$. For LLama-3-8B-Instruct, we resample at every token, whereas for DeepSeekR1-Distill-Llama-8B we resample at every sentence, since reasoning models generate much longer reasoning chains. In total, we collected $1.77$B tokens for this FPA dataset.

We evaluate forking path analyses with decreasing sampling rates $S \in [5,100]$ and intervals $N \in \{1, 2, 4, 8, 16, 32\}$, against the reference analysis with $S{=}200$ and $N={1}$. We report total variation distance (TVD) between this reference FPA against each lower-sample analysis to characterize the tradeoff between precision and sampling efficiency.
For each of these conditions, we then use our statistical model to produced smoothed $o_t$ estimates, and compare these to the $S{=}200$ data.

\subsection{Estimating $o_t$ from fewer samples}
\label{sec:results-recovery}

Across all questions, the statistical model's excess TVD above the reference noise floor falls monotonically from $0.0265$ ($S{=}5$) to $0.0056$ ($S{=}100$). For Llama, the sample efficiency multiplier at 4-token spacing is $5.0\times$ at $S{=}5$, $3.3\times$ at $S{=}30$, and $1.0\times$ at $S{=}100$ (Figure~\ref{fig:accuracy}, dark blue lines). At 1-token spacing, the sample efficiency multiplier grows to $22.1\times$ at $S{=}5$, $7.3\times$ at $S{=}30$, and $1.6\times$ at $S{=}100$ (Figure~\ref{fig:accuracy}, red lines). We find similar results with the reasoning model's sentence resampling with multipliers $4.9\times$ at $S{=}5$, $3.1\times$ at $S{=}30$, and $1.1\times$ at $S{=}100$ (Appendix~\ref{app:deepseek}).

We find that with raw data, increasing the spacing between observation has relatively little effect on accuracy overall (Fig.~\ref{fig:accuracy}, dotted lines), except with high $S$. This may be because sampling noise dominates fluctuations in $o_t$ unless $S$ is sufficiently high (e.g., Fig.~\ref{fig:reconstruction}, Bottom). An exception to this trend is regions around forking points, where using a wider stride gives less accurate $o_t$ estimates (Fig.~\ref{fig:fork-accuracy}).
However, the smoothed data shows a very different pattern, where higher sampling granularity reliably leads to more accurate $o_t$ estimates (Fig.~\ref{fig:accuracy}, solid lines), which also holds for forking regions (Fig.~\ref{fig:fork-accuracy}).



\begin{figure}[t]
\centering
\includegraphics[width=\columnwidth]{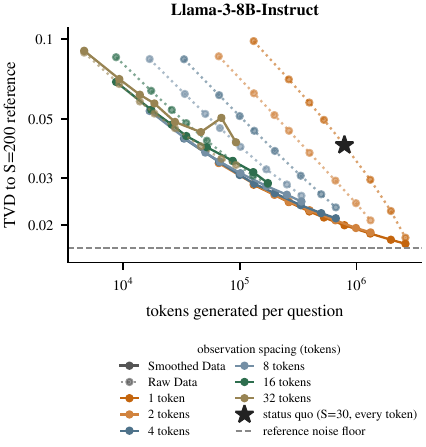}
\caption{\textbf{Accuracy versus token cost.} \ \ 
We find that overall, with raw $o_t$ data, wider spacing between resampled points $N$ leads to better accuracy with lower overall token cost (dotted lines). Our statistical model significantly increases the accuracy for most $S$ and $N$, but also negates the improvement of larger $N$ increasing accuracy with equal cost (solid lines).
TVD between the $S{=}200$ reference and each low-sample estimator is shown on the y-axis, with colors representing different $N$. 
The black star represents the sampling parameters used in \citep{bigelow2024forking}.
}
\label{fig:budget}
\end{figure}

\subsection{Tradeoff between accuracy and cost}
\label{sec:results-power}

Figure~\ref{fig:budget} maps accuracy against token cost, comparing the raw data against model-smoothed data. 
Here, we see a clear pattern with the raw data: wider token strides lead to more accurate approximations of high-sample data, given the same token budget. Intuitively, averaging data over a fixed window leads to the $o_t$ estimates having lower variance, since sampling noise is averaged out.
When data is smoothed, however, we instead see that changing sampling window has relatively little effect on token efficiency, but instead represents a different section of the same Pareto frontier. In other words, with smoothed data, changing $S$ and the spacing between observation samples $t$ has roughly the same effect (Fig.~\ref{fig:budget}, solid lines).

\subsection{Sampling noise decreases with sample size}
\label{sec:results-noise}

To measure how sampling noise changes with sample size, we divided the $S{=}1000$ samples for each of two questions into disjoint, equally sized subsets. At each position, we then computed the pairwise total variation distance (TVD) between the outcome distributions estimated from these subsets.
As Figure~\ref{fig:noise-law} shows, the pooled TVD between estimates from these independent sample subsets decreases steadily as $S$ increases. On log-log axes, the fitted slope is $-0.4903$ through $S{=}200$, closely matching the $-1/2$ slope expected under independent multinomial sampling, suggesting that noise decreases in proportion to $1/\sqrt{S}$. 

The full $S{=}200$ dataset shows similar scaling, with slopes of $-0.4757$ for Llama when resampling every token and $-0.4838$ for DeepSeek when resampling every sentence. At $S{=}200$, the variation across runs ($\text{TVD}=0.0451$) is nearly identical to the sampling variation within a run ($\text{TVD}=0.0442$). These results support modeling the observed counts at each position as multinomial samples: $c_t \sim \mathrm{Mult}(S, o_t)$.


\section{Discussion}
\label{sec:discussion}

Here we have shown that heavy cost of resampling steps in reasoning and text generation is in fact a modeling problem. The noise of $o_t$ is exactly multinomial and the curves are smooth with sparse forking points, so a large resampling budget mostly re-measures structure that neighboring positions already contain. An estimator that pools within segments and breaks at change points converts that structure into $\sim15\times$ efficiency gains within segments, and $\sim4\times$ efficiency gains without losing accuracy in identifying forking points.

This work opens a number of exciting avenues for future work. It suggests that there could be a mechanistic theory of in-context learning that can explain these kinds of learning dynamics \citep{nanda2023progress, bigelow2024context}. Finally, while our method enables us to post-hoc smooth approximate outcome distributions $o_t$, an alternate approach which integrates modeling would be to more efficiently choose particular token indices to sample from, similar to approaches in optimal experiment design \citep{chaloner1995bayesian}.


\bibliography{refs}

\newpage
\appendix

\section{Limitations}

\textbf{Scope.} Our experiments are limited to tinyMMLU multiple-choice questions and two 8B-parameter models: one instruction-tuned model and one reasoning-distilled model. We analyze five-class outcome distributions over relatively short rollouts (continuations of at most 400 tokens for Llama and 1536 tokens for DeepSeek). Whether the results generalize to longer horizons, open-ended outcomes, or larger models remains to be tested.

\textbf{Hyperparameter tuning.} The Full Model and its hyperparameter tuning procedure were developed using two questions from one model. The tinyMMLU $S{=}200$ dataset and all DeepSeek data were used only for evaluation; however, they come from the same task family and use models at the same parameter scale as the development data.

\textbf{Estimator effectiveness depends on how big the forks are.} The estimator's advantage in fork regions is established at a forking threshold of $0.10$, but it degrades at $0.15$ ($n{=}35$ questions), and at the largest forks of $0.20$ ($n{=}21$), smoothing is a marginally worse estimator than raw data.

\textbf{Point estimates only.} The estimators' 90\% credible intervals do not achieve their nominal coverage on the observed data: the Full Model's empirical coverage ranges from $0.48$ to $0.64$ across the development questions. We therefore report and recommend the point estimates rather than the credible intervals.

\textbf{Reference curves are estimates.} The reference curves are themselves finite-sample estimates ($S{=}200$ or $S{=}1000$) and thus have a nonzero noise floor. Figure-level TVDs computed against the full reference include this floor. For hypothesis-bearing comparisons, we use leave-replicate-out references to prevent overlap between an estimate and its reference from artificially reducing the measured error.

\section{Additional Example Reconstructions}
\label{app:more-recons}

\begin{figure*}[h]
\centering
\includegraphics[width=\textwidth]{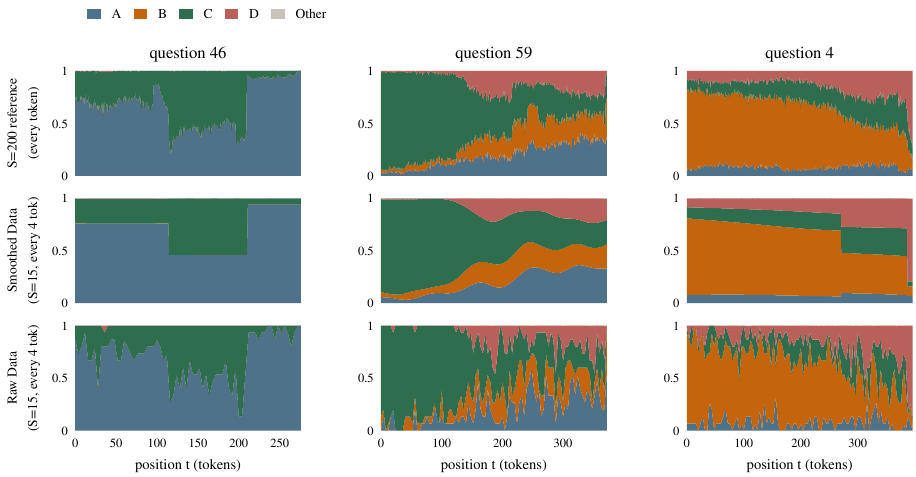}
\caption{Reconstructions for three questions from the full tinyMMLU evaluation set using Llama-3-8B-Instruct. From top to bottom, the rows show the $S{=}200$ every-token reference, the Full Model's reconstruction from a nested $S{=}15$ run at 4-token spacing, and Raw Data from the same reduced run.}
\label{fig:more-recons}
\end{figure*}

Figure~\ref{fig:more-recons} extends the comparison in Figure~\ref{fig:reconstruction} to three questions from the tinyMMLU $S{=}200$ evaluation set. For each question, the reference is the $S{=}200$ every-token curve. The reduced run uses $S{=}15$ at 4-token spacing, with samples nested within the same set of draws.

\section{DeepSeek-R1-Distill Results}
\label{app:deepseek}

\begin{figure}[h]
\centering
\includegraphics[width=\columnwidth]{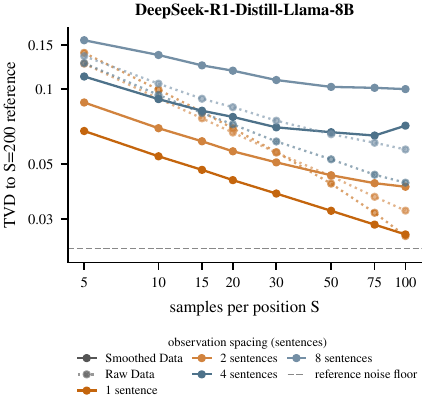}
\caption{DeepSeek counterpart to Figure~\ref{fig:accuracy}. Pooled TVD to the full $S{=}200$ reference is plotted against the number of samples per position, $S$, with one line per observation spacing in sentences. Dotted lines are results for raw data, solid lines are smoothed data.}
\label{fig:dseek-accuracy}
\end{figure}

\begin{figure}[h]
\centering
\includegraphics[width=\columnwidth]{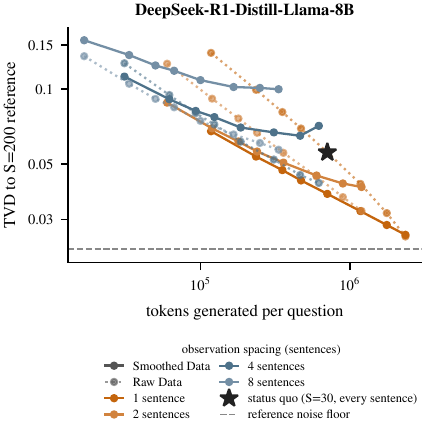}
\caption{DeepSeek counterpart to Figure~\ref{fig:budget}. Pooled TVD is plotted against the number of tokens generated per question; colors match Figure~\ref{fig:dseek-accuracy}. Dotted lines are results for raw data, solid lines are smoothed data. The star marks the baseline setting of $S{=}30$ at 1-sentence spacing.}
\label{fig:dseek-budget}
\end{figure}

Figures~\ref{fig:dseek-accuracy} and~\ref{fig:dseek-budget} reproduce the analyses in main-text Figures~\ref{fig:accuracy} and~\ref{fig:budget} for DeepSeek-R1-Distill-Llama-8B. They show accuracy and sampling cost on the reasoning model with sentence-level resampling, with observation spacings from $N{=}1$ to $N{=}8$ sentences. The main trends carry over: TVD decreases as $S$ increases at every spacing, the densest observation spacing performs best, and the Full Model yields its largest improvement over Raw Data at small $S$. Section~\ref{sec:results-recovery} reports the corresponding effective-sample multipliers.

\section{Forking Region Accuracy and Cost}
\label{app:fork}

\begin{figure*}[h]
\centering
\includegraphics[width=\textwidth]{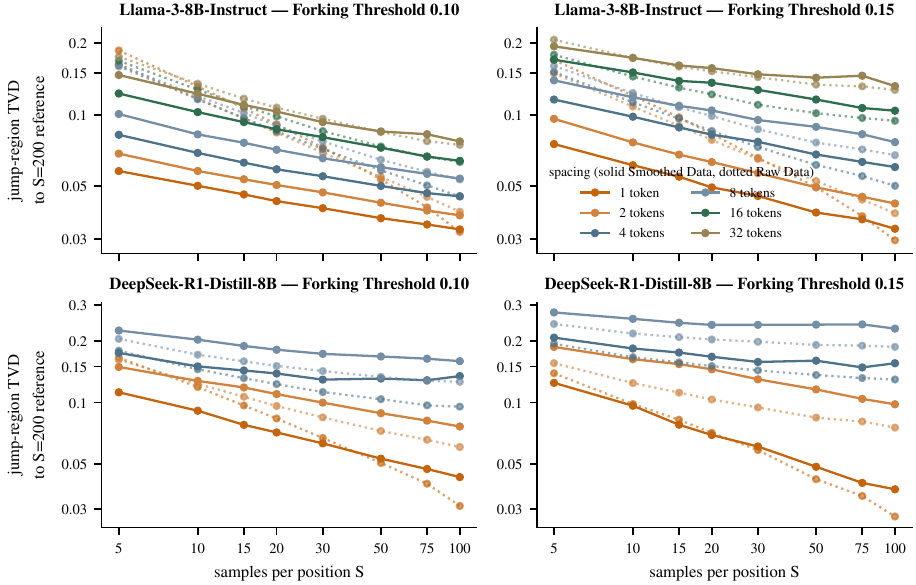}
\caption{Forking region counterpart to Figure~\ref{fig:accuracy}. TVD to the $S{=}200$ reference is evaluated only at forking regions, using forking thresholds of $0.10$ (left) and $0.15$ (right). Solid lines show Smoothed Data, and dotted lines show Raw Data; colors match those in the main text.}
\label{fig:fork-accuracy}
\end{figure*}

\begin{figure*}[h]
\centering
\includegraphics[width=\textwidth]{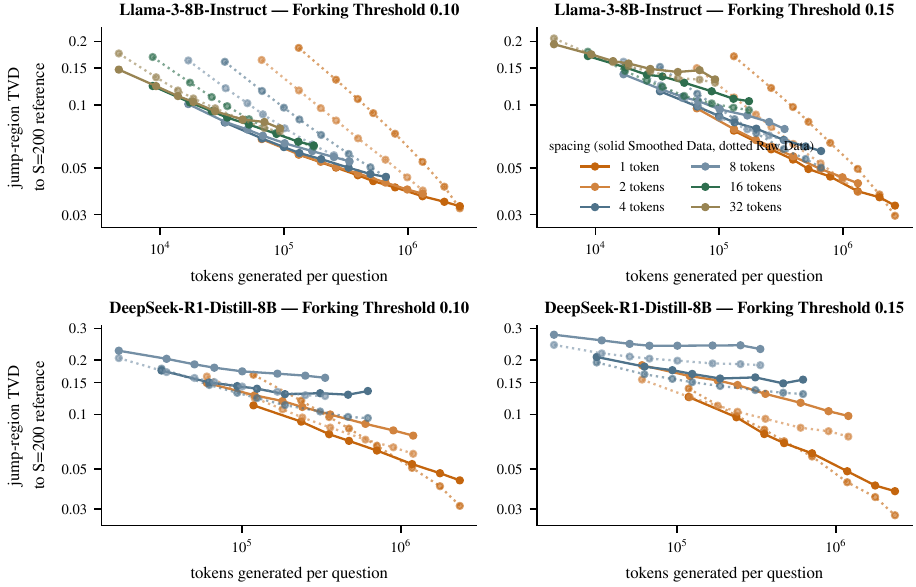}
\caption{Fork-region counterpart to Figure~\ref{fig:budget}. TVD in forking regions is plotted against the number of tokens generated per question at forking thresholds of $0.10$ and $0.15$. Solid lines show Smoothed Data, and dotted lines show Raw Data.}
\label{fig:fork-budget}
\end{figure*}

Figures~\ref{fig:fork-accuracy} and~\ref{fig:fork-budget} reproduce the accuracy and budget analyses from the main text after restricting evaluation to forking regions, where a forking region is defined as positions within $\pm 10$ tokens of a forking point. Forking points are defined as points $t$ such that $TVD(o_t, o_{t+N}) > \epsilon$, i.e. points where there is a significant difference in the outcome distribution between two adjacent steps. We report results at forking thresholds $\epsilon \in \{0.10, 0.15, 0.20\}$.

We observer that performance in forking regions deteriorates earlier and more rapidly than overall performance as the observation spacing becomes coarser and $N$ increases. At $S{=}30$, fork-region TVD increases from $0.0669$ to $0.0936$ to $0.1587$ as spacing increases from $N{=}4$ to $N{=}8$ and $N{=}16$ tokens. It then saturates near $0.16$ at spacings of 32--64 tokens, indicating that the fork has effectively been missed. Flat-region performance degrades more gradually, and the same ordering holds at scale for both models.

Second, the Full Model's advantage over Raw Data near forks is concentrated at moderate forks (threshold $0.10$), disappears for larger forks, and even reverses with the largest forks (Section~\ref{sec:results-power}). In the threshold-$0.15$ panels, the solid and dotted curves at dense spacings nearly coincide. This agreement reflects the powered null result of the preregistered test rather than a plotting artifact.

\section{Component Ablation of the Full Model}
\label{app:ablation}

Ablating the Full Model's components one at a time on the $S{=}200$ tinyMMLU dataset reveals which components contribute to accuracy in flat and forking regions. We report paired per-question TVD differences relative to the Full Model, computed against leave-replicate-out references, with 95\% $t$-intervals.

\subsection{Estimators}

We compare four estimators of increasing complexity:
\begin{description}\itemsep2pt\parskip0pt
\item[Raw Data:] Empirical frequencies with linear interpolation between observed positions.
\item[Kernel Pooling:] Gaussian-kernel-weighted Dirichlet pooling of neighboring counts. This estimator is designed for regions between forks, where increments are small and nearby positions have similar values of $o_t$.
\item[Segment+Pool:] PELT change point detection
\citep{killick2012optimal,truong2020selective} with an exact multinomial cost, followed by per-segment Dirichlet pooling. The resulting piecewise-constant fit preserves boundaries at detected forks.
\item[Full Model:] PELT segmentation followed by Kernel Pooling truncated at change point boundaries. This estimator accommodates gradual drift within segments while preserving sharp changes at forks. The penalty and bandwidth hyperparameters are selected by cross-validation on low-sample runs.
\end{description}


\subsection{Further Model Ablations}

\begin{figure*}[h]
\centering
\includegraphics[width=\textwidth]{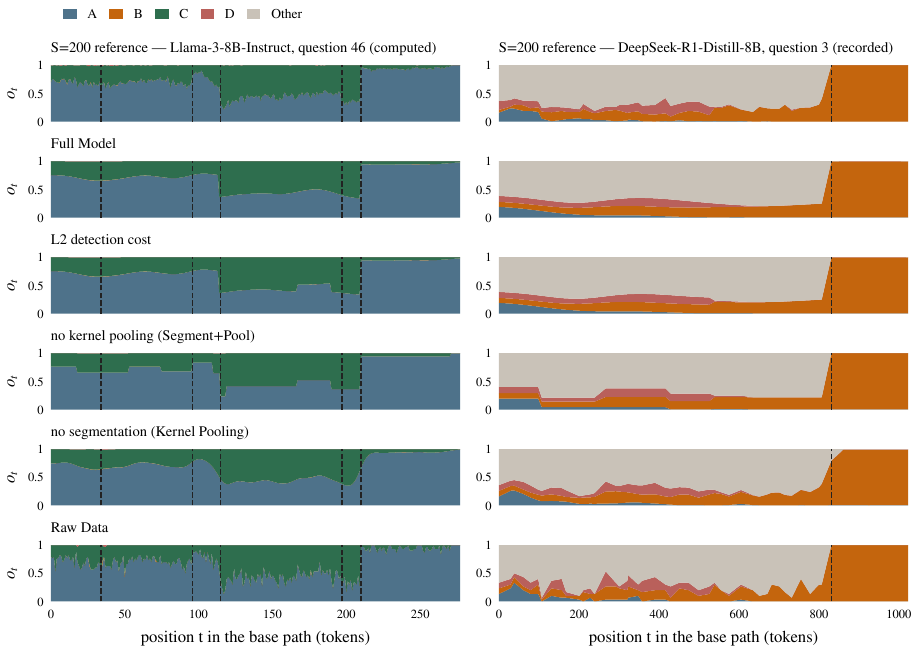}
\caption{Ablation overlays for two example questions. Reconstructions use $S{=}30$ on the densest spacing $N$ and are evaluated against leave-replicate-out $S{=}200$ references; dashed lines mark reference forks. The left column shows Llama-3-8B-Instruct question~46 using the ablation variants, cross-validation, and operating point from the recorded $S{=}200$ store. The right column shows the recorded DeepSeek example, where the fork occurs when continuations stop resolving to an answer. Removing segmentation (Kernel Pooling) blurs the forks, whereas removing kernel pooling (Segment+Pool) reduces gradual drift to discrete steps. The L2 detection-cost variant is visually indistinguishable from the Full Model.}
\label{fig:ablation-overlay}
\end{figure*}

\textbf{Segmentation preserves forks.} Removing segmentation and using Kernel Pooling alone retains 93--96\% of the pooled improvement but degrades accuracy in forking regions. The TVD difference relative to the Full Model is $+0.0113$ $[+0.0039, +0.0187]$ for Llama at forking threshold $0.15$ and $+0.0091$ $[+0.0036, +0.0145]$ for DeepSeek at threshold $0.10$. Figure~\ref{fig:ablation-overlay} shows two examples in which Kernel Pooling blurs forks that the Full Model preserves. On the recorded DeepSeek example, the respective TVDs are $0.2720$ and $0.0088$.

\textbf{Cross-validated tuning matters at large forks.} Replacing cross-validated hyperparameters with fixed values increases TVD by $0.0126$ $[+0.0051, +0.0202]$ for Llama at threshold $0.15$.

\textbf{Kernel pooling provides a small but consistent pooled gain.} Replacing kernel pooling with flat per-segment pooling (Segment+Pool) increases pooled TVD by $0.0021$ $[+0.0015, +0.0026]$ for Llama at $S{=}30$.

\textbf{The detection cost has little end-to-end effect.} Fixing PELT to the L2 cost increases pooled TVD by $\leq 0.0010$ throughout the ablation. Its performance is statistically indistinguishable from that of the Full Model in both headline fork conditions; for example, the difference for Llama at $S{=}30$ and threshold $0.10$ is $+0.0003$ $[-0.0004, +0.0009]$. This variant retains 97--100\% of the Full Model's improvement over Raw Data.

\begin{figure*}[h]
\centering
\includegraphics[width=0.82\textwidth]{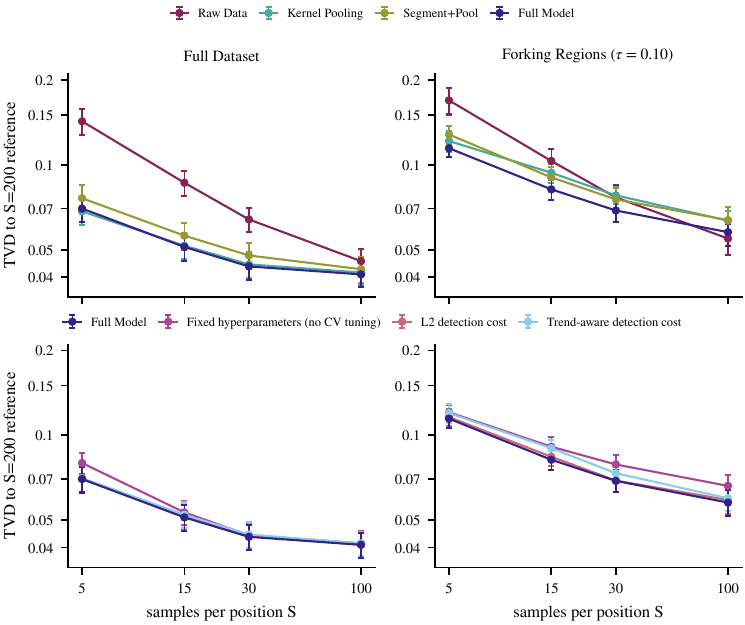}
\caption{Paired per-question TVD differences between each ablation variant and the Full Model for DeepSeek-R1-Distill-Llama-8B at 1-sentence spacing. Error bars show 95\% $t$-intervals over tinyMMLU, and the dashed zero line denotes the Full Model. Results are shown for pooled accuracy (left) and forking region accuracy at threshold $0.10$ (right). Removing segmentation harms accuracy near forks but not pooled accuracy; fixing the detection cost to L2 is indistinguishable from the Full Model in both panels.}
\label{fig:ablation-deltas}
\end{figure*}

Figure~\ref{fig:ablation-deltas} summarizes these paired differences. For pooled accuracy, every smoothing variant is within a few thousandths of the Full Model, whereas Raw Data has substantially higher TVD. In forking regions, the no-segmentation variant differs reliably from zero, while the L2-detection variant does not.

\section{The $1/\sqrt{S}$ Noise Law}
\label{app:noise}

\begin{figure}[h]
\centering
\includegraphics[width=\columnwidth]{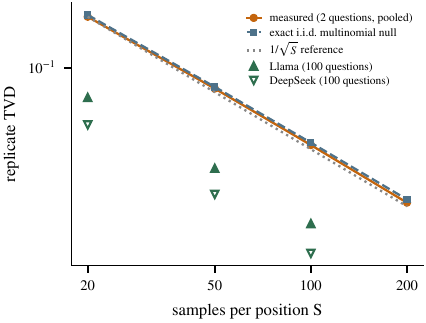}
\caption{Replicate TVD versus the number of samples per position, $S$, on log--log axes. The measured curve remains within $2\%$ of the exact i.i.d.\ multinomial null through $S{=}200$. Triangles show the full dataset means for both models.}
\label{fig:noise-law}
\end{figure}

Figure~\ref{fig:noise-law} presents the replicate-noise measurements underlying Section~\ref{sec:results-noise}. For the $S{=}1000$ development set, pooled TVD between disjoint replicates is plotted against $S$ on log--log axes. The fitted slope is $-0.4903$ overall and $-0.5047$ in the tail. The figure also shows the exact i.i.d.\ multinomial null, computed by resampling the per-branch histograms through the same statistic; the ratios of measured to null TVD range from $0.98$ to $1.01$. Finally, the per-model means from the $S{=}200$ tinyMMLU dataset have slopes of $-0.4767$ and $-0.4577$. Their absolute TVDs are lower because those questions are less variable on average, but the error decreases at a similar rate.

\section{Segmentation Cost in PELT Implementation}
\label{app:segmentation}

During this work, we uncovered a bug in the implementation of PELT in the \texttt{ruptures} library \citep{truong2020selective}.

\begin{figure*}[t]
\centering
\includegraphics[width=0.92\textwidth]{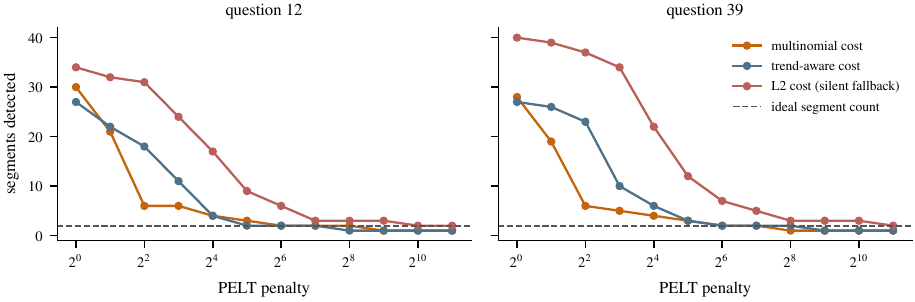}
\caption{Number of detected segments as a function of the PELT penalty
for $S{=}30$ dense-grid counts on the two development questions. Curves
show the exact multinomial cost, a trend-aware cost, and the silently
substituted L2 cost. The dashed line marks the ideal segment count.}
\label{fig:seg-count}
\end{figure*}

The segment-based estimators use PELT change point detection with an exact multinomial cost. If a custom cost object does not inherit from \texttt{BaseCost}, the library silently replaces it with the default least-squares cost, \texttt{CostL2}. The fit still completes and returns change points without warning, but results are significantly affected (Figure~\ref{fig:seg-count}).



\section{LLM Use Statement}
\label{app:silico-use}

The research reported in this paper and the initial manuscript draft were produced by the autonomous LLM agent Silico (\url{https://www.goodfire.com/silico}) under human direction.

\end{document}